\documentclass[letterpaper]{article} % DO NOT CHANGE THIS
\usepackage{aaai2027}  % DO NOT CHANGE THIS
\nocopyright
\usepackage[hyphens]{url}  % DO NOT CHANGE THIS
\usepackage{graphicx} % DO NOT CHANGE THIS
\usepackage{natbib}  % DO NOT CHANGE THIS AND DO NOT ADD ANY OPTIONS TO IT
\usepackage{caption} % DO NOT CHANGE THIS AND DO NOT ADD ANY OPTIONS TO IT
\usepackage{algorithm}
\usepackage{algorithmic}
\usepackage{amsmath}
\usepackage{amssymb}

\usepackage{newfloat}
\usepackage{listings}
\DeclareCaptionStyle{ruled}{labelfont=normalfont,labelsep=colon,strut=off} % DO NOT CHANGE THIS
\floatstyle{ruled}
\newfloat{listing}{tb}{lst}{}
\floatname{listing}{Listing}

\usepackage{booktabs}
\usepackage{pifont}
\newcommand{\cmark}{\ding{51}}
\newcommand{\xmark}{\ding{55}}
\usepackage[table]{xcolor}

\title{SkillTFM: Gated Skill Evolution for \\
Training-Free Adaptation of Tabular Foundation Models}

\author{
    Yi He\textsuperscript{\rm 1},
    Zhengkang Guan\textsuperscript{\rm 1},
    Anpeng Wu\textsuperscript{\rm 1},
    Peng Cui\textsuperscript{\rm 2},
    Fei Wu\textsuperscript{\rm 1},
    Kun Kuang\textsuperscript{\rm 1}\corresponding
}

\affiliations{
    \textsuperscript{\rm 1}Zhejiang University, Hangzhou, China\\
    \textsuperscript{\rm 2}Tsinghua University, Beijing, China\\
    12521297@zju.edu.cn,
    kunkuang@zju.edu.cn
}

\begin{document}

\maketitle

\begin{abstract}
Tabular data are ubiquitous in real-world applications and are crucial for data-driven prediction and decision-making across science, industry, finance, healthcare, and public services. Tabular foundation models (TFMs) have emerged as a promising paradigm for general-purpose tabular learning, offering reusable predictors across diverse datasets and substantially reducing the need for task-specific training, tuning, and model development. However, their practical deployment remains constrained by distribution shifts, heterogeneous feature semantics, and task-specific patterns that are difficult to capture without costly fine-tuning or additional labeled data.

To this end, we propose \textbf{SkillTFM}, a training-free system that shifts TFM adaptation from parameter updates to the gated evolution of agentic skills. The core of SkillTFM is a verifiable and extensible skill bank that couples boundary evidence identification with gated skill evolution: the former characterizes task structure and base-model failure patterns, whereas the latter retrieves and extends reusable skills subject to explicit validation. Across simulated boundary settings and real-world electricity-price forecasting, SkillTFM improves AUC by $0.128$--$0.142$, raises nonlinear-boundary AUC from $0.699$ to $0.898$. Furthermore, experiments across TFM backbones demonstrate the effectiveness and generality of SkillTFM.
\end{abstract}

\section{Introduction}
\begin{figure*}[t]
    \centering
    \includegraphics[width=0.85\linewidth]{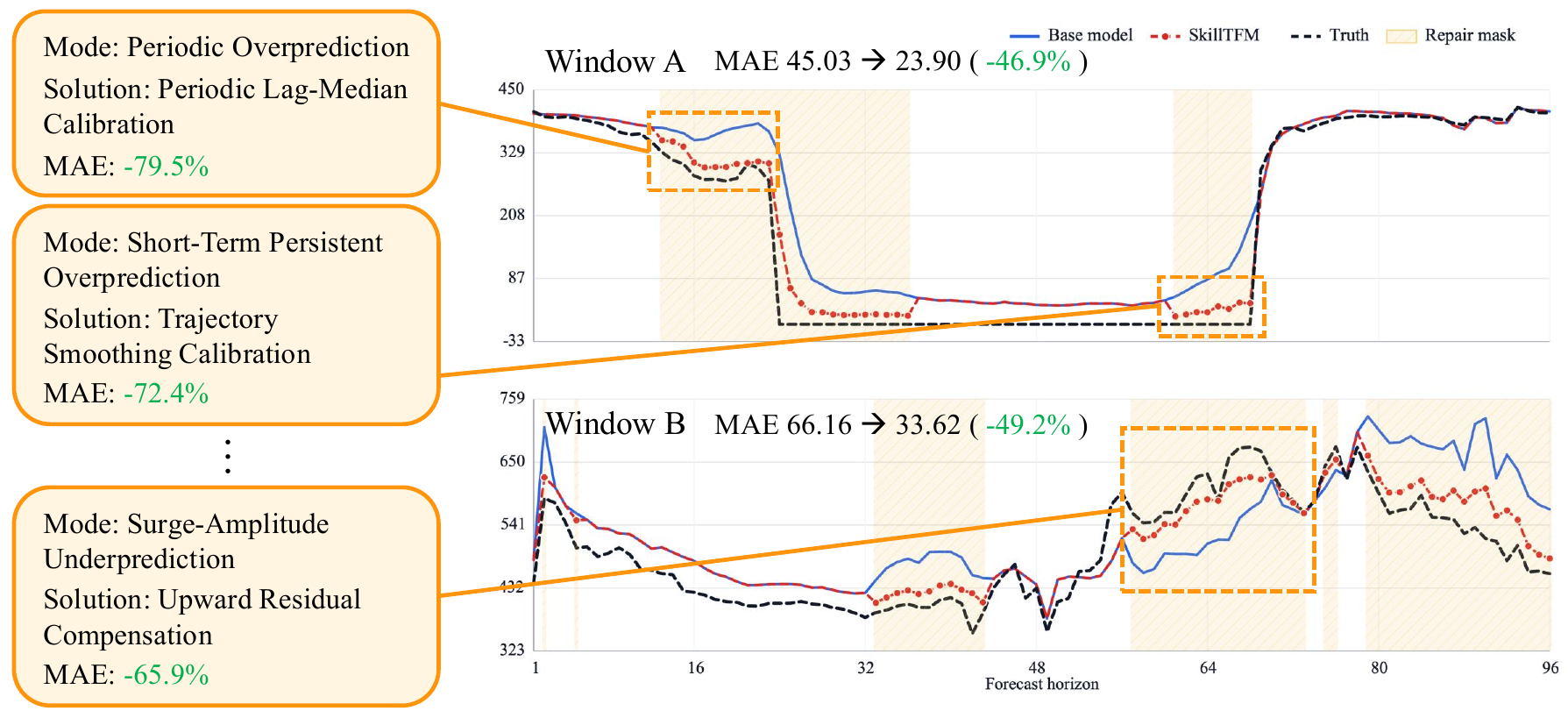}
    \caption{Selective repair in electricity price forecasting windows from two cities. The annotated regions show how SkillTFM converts boundary evidence into repairs: periodic overestimation, persistent overestimation while prices remain low, and underestimation during an upward ramp. Evidence-supported repairs reduce MAE from 45.03 to 23.90 and from 66.16 to 33.62; otherwise, SkillTFM preserves the base-model trajectory through fallback.
    }
    \label{fig:electricity-repair}
\end{figure*}

Tabular data are widely used in science, finance, industry, and electricity systems, but conventional tabular modeling requires dataset-specific model selection, tuning, and preprocessing \cite{wang2026limix2mmitigatinglowrankcollapse}. Tabular foundation models provide an attractive alternative: a general-purpose predictor can serve tabular tasks through task context \cite{hollmann2023tabpfntransformersolvessmall,Hollmann2025,qu2025tabicltabularfoundationmodel}. This paradigm reduces task-specific training and tuning, allowing the same base model to act as a unified prediction interface across datasets. However, it also raises a new question: when a task falls outside the structural conditions that the model already handles well, how can the system expand its usable boundary?

Directly using the base TFM is efficient, but under unstable dependencies, distribution shifts, missingness, nonlinear structure, or systematic errors in deployment, it may preserve clear boundary failures \cite{Geirhos2020,10.1016/j.patcog.2011.06.019,pmlr-v139-koh21a,JMLR:v8:sugiyama07a,3305381.3305518}. Another natural approach is to add fixed preprocessing, rule-based repair, calibration, or a manually designed repair pipeline. However, such methods often treat repair as the default action: they may improve some failures, but they can also intervene on tasks where the base model is already reliable. Therefore, expanding the boundary of TFMs is not simply a matter of replacing the model or writing one rule for each failure type \cite{JMLR:v11:el-yaniv10a,pmlr-v97-geifman19a}; it requires deciding whether the current task provides enough evidence to support repair, and whether a proposed skill update can pass explicit validation \cite{yang2026skilloptexecutivestrategyselfevolving,li2026skillsbenchbenchmarkingagentskills,zhang2026coevoskillsselfevolvingagentskills}.

To address this problem, we propose \textbf{SkillTFM}, a training-free adaptation system that equips tabular foundation models with a gated external skill state. SkillTFM extracts boundary evidence from the current task, including feature distribution, label-feature relation, prediction bias, trend/periodicity, missingness pattern, and probe sensitivity. This evidence activates verified skills in a repair skill bank and forms candidate repairs. Candidate repairs are then ranked and checked by a runtime certificate; a repair is executed only when evidence is sufficient, risk is acceptable, and historical memory does not block the route. If no candidate repair is certified, SkillTFM falls back to the base prediction. SkillTFM further expands its skill state through validation-gated skill evolution: candidate edits can become deployed skills only after schema checking, selection improvement, no-harm validation, and regression review.

Figure~\ref{fig:electricity-repair} illustrates how SkillTFM applies this principle in real electricity-price forecasting windows. The base model does not fail globally, but exhibits systematic boundary errors in local regions: in Window A, it shows periodic overprediction and short-term persistent overprediction; in Window B, it underestimates the amplitude of rapid upward ramps. SkillTFM executes repairs in regions supported by local evidence, moving the prediction trajectory closer to the realized price. In regions where evidence is insufficient or the repair is not certified, SkillTFM preserves the base-model trajectory rather than forcing a modification. Thus, TFM boundary failure is often not a global switch, but depends on local data patterns, base-model error trends, and repair risk. Through such selective repair, MAE is reduced from 45.03 to 23.90 and from 66.16 to 33.62 in the two windows, respectively.

We evaluate SkillTFM on structured boundary shifts, mixed-boundary settings, generator/severity transfer, nonlinear skill evolution, cross-model transfer, cross-optimizer portability, ablations, and real-world electricity-price forecasting. On held-out boundary evaluations, SkillTFM improves AUC by 0.128--0.142 with zero observed harm. On an unsupported nonlinear boundary, skill evolution improves AUC from 0.699 to 0.898. The skill state transfers across TFM backbones and remains controlled by the local promotion gate under different optimizer proposers. In electricity-price forecasting, SkillTFM reduces MAE while modifying only part of the prediction trajectory, further demonstrating selective intervention. Overall, our contributions are: (1) to our knowledge, we propose the first skill-based adaptation system for training-free tabular foundation models, and show that this pluggable skill system can be attached to different agents and TFMs; (2) a gated skill-evolution mechanism that couples selective repair with safe fallback; and (3) evidence that the learned skill state transfers across boundary settings, TFM backbones, and optimizer proposers.

\section{Related Work}

\paragraph{Tabular Foundation Models and Boundary Repair.}
Tabular foundation models replace per-dataset training with reusable predictors operating from task context, following prior-data fitted networks, in-context tabular learners, and hypernetwork-based predictors \cite{DBLP:journals/corr/abs-2112-10510,pmlr-v202-nagler23a,muller2025mothernetfasttraininginference}. Recent work further studies scalable architectures, structure-aware modeling, transfer or real-data pretraining, stronger TabPFN variants, and stability or causality-oriented foundation models \cite{ma2026tabdptscalingtabularfoundation,zhang2025limixunleashingstructureddatamodeling,kim2024cartepretrainingtransfertabular,garg2025realtabpfnimprovingtabularfoundation,grinsztajn2026tabpfn25advancingstateart,guan2026stablepfn,guan2026dcdpfn,chen2026dagfm}. Robust tabular learning and uncertainty methods address imbalance, missingness, label noise, calibration, out-of-distribution detection, abstention, and calibrated prediction sets \cite{5128907,Emmanuel2021,Frenay2014,10.1093/biomet/63.3.581,NEURIPS2019_8558cb40,angelopoulos2022gentleintroductionconformalprediction,JMLR:v11:el-yaniv10a}. These works address important pieces of tabular prediction and robustness, but they usually introduce new backbones, single mechanisms, or fixed pipelines. SkillTFM instead learns an external skill state that treats repair as an evidence-conditioned action: the system may execute a repair, fall back, or reject a candidate based on task evidence, risk, and memory of prior failures \cite{pmlr-v97-geifman19a,hendrycks2018baselinedetectingmisclassifiedoutofdistribution}.

\paragraph{Skill Learning and Validation-Gated Updates.}
Recent work on agent skills studies skill repositories, memory-based reuse, self-evolving agents, and empirical analyses of agent skill learning \cite{jiang2026sokagenticskills,ouyang2026skilloslearningskillcuration,shen2026skillfoundrybuildingselfevolvingagent,zhang2026memskilllearningevolvingmemory,ling2026agentskillsdatadrivenanalysis,zhou2026comprehensivesurveyagentskills}. Related agent systems use tools, reflection, and experiential memory \cite{wang2023voyageropenendedembodiedagent,shinn2023reflexionlanguageagentsverbal,zhao2024expelllmagentsexperiential,yao2023reactsynergizingreasoningacting,schick2023toolformerlanguagemodelsteach}, while prompt and program optimization treats textual or modular components as revisable objects \cite{khattab2023dspycompilingdeclarativelanguage,yuksekgonul2024textgradautomaticdifferentiationtext,yang2024largelanguagemodelsoptimizers}. Reflective prompt-evolution work further highlights the need to validate generated edits before deployment \cite{agrawal2026gepareflectivepromptevolution}. Existing language-agent or prompt-level skill methods are usually tied to specific LLMs and domain problems, with limited transferability across models and problem settings; therefore, they are difficult to apply to structured boundary failures in TFM deployment. SkillTFM instead brings skill-based adaptation to tabular foundation models and connects to different TFM backbones and LLM-based skill proposers in a pluggable manner; it maintains an external tabular repair state through gated updates, admitting candidate skills only after explicit validation.

\begin{figure*}[t]
    \centering
    \includegraphics[width=\linewidth]{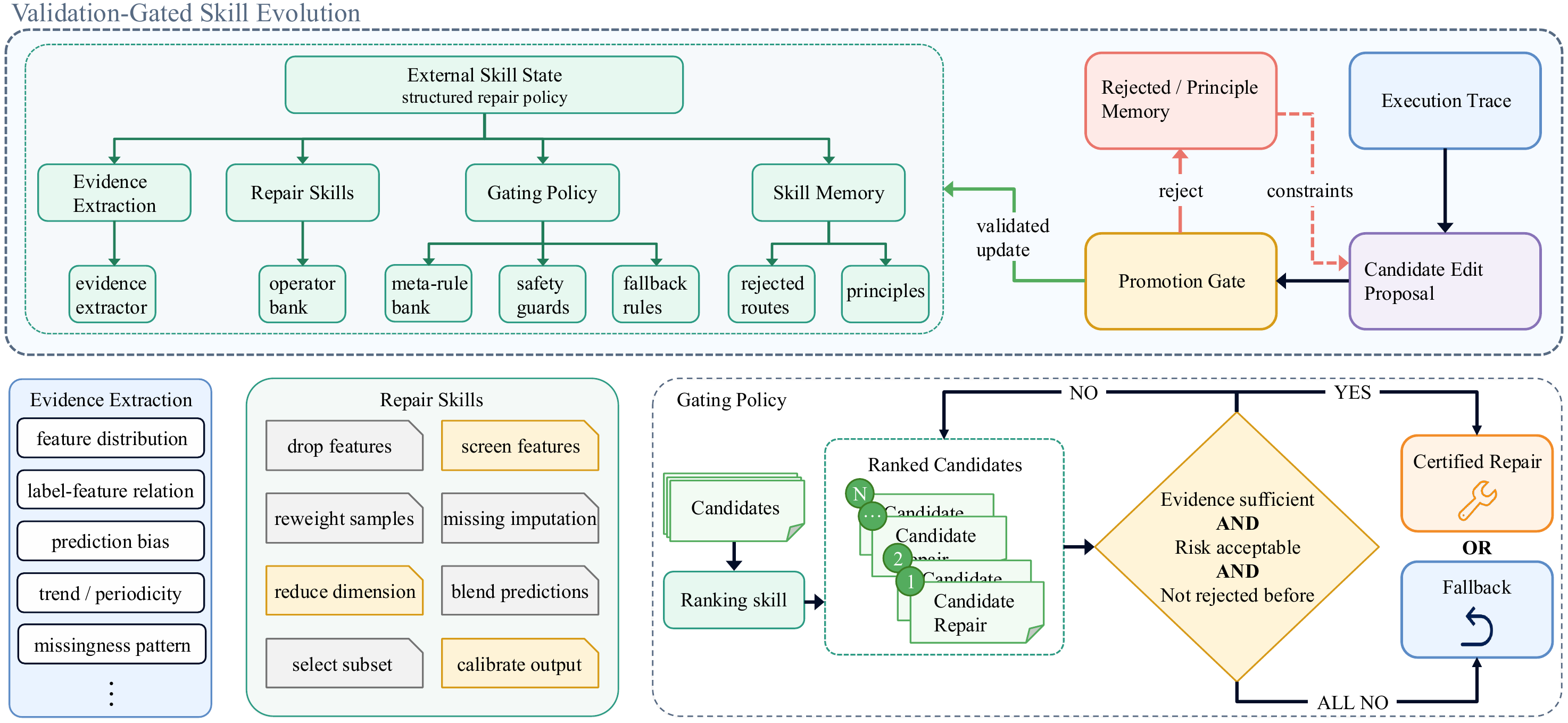}
    \caption{Overview of SkillTFM. The lower panels show evidence-conditioned boundary expansion, where task evidence activates verified repair skills and the gating policy selects either a certified repair or fallback; yellow highlights indicate skills or routes activated by the current evidence. The upper loop shows validation-gated skill evolution, where execution traces generate candidate edits, accepted edits update the external skill state, and rejected edits become memory constraints for future decisions.
    }
    \label{fig:methodology}
\end{figure*}

\section{Methodology}

SkillTFM formulates training-free adaptation of tabular foundation models as an evidence-conditioned boundary expansion decision. Given a tabular task, the base model first produces an original prediction; SkillTFM then decides whether the current task provides sufficient evidence for safely expanding the model's usable boundary through an external skill state. If the current evidence supports a verified skill, the system executes a certified repair; if the evidence is insufficient, the risk is high, or historical memory constrains a similar route, the system falls back to the base-model prediction. Figure~\ref{fig:methodology} illustrates this process: the lower part shows how the skill state operates on the current task through evidence extraction, repair skill activation, and the gating policy; the upper part shows how execution traces generate candidate edits, which either become validated updates through the promotion gate or enter memory as constraints on future evolution.

\subsection{Problem Setup: Boundary Expansion Decision}

Let \(M\) denote a tabular foundation model. For a tabular task \(T\), the labeled training context is
\begin{equation}
D_{\mathrm{train}}=(X_{\mathrm{train}},y_{\mathrm{train}}),
\end{equation}
and the observable deployment context includes unlabeled covariates \(X_{\mathrm{test}}\). The base model conditions on \(D_{\mathrm{train}}\) and \(X_{\mathrm{test}}\), producing an original prediction \(\hat{y}_M\) or probability estimate \(\hat{p}_M\).

The core problem in SkillTFM is not simply whether to modify a prediction point, but whether the current task lies in a boundary region that can be safely covered by an external skill. A task may be in one of three states: the base model is already reliable; the base model exhibits a repairable boundary failure; or the current evidence and risk are insufficient to support safe expansion. Certified repair and fallback are therefore two execution outcomes of the boundary expansion decision: the former indicates that the current boundary can be expanded, while the latter preserves the base-model output.

Let \(S\) denote the external skill state, and let \(S_{\mathrm{base}}\) denote the fallback-only reference state that always returns the base-model output. SkillTFM aims to improve task utility while controlling harmful intervention during boundary expansion:
\begin{equation}
\max_S \mathbb{E}_{T\sim\mathcal{D}}
\left[U(T,S)-U(T,S_{\mathrm{base}})\right],
\end{equation}
subject to
\begin{equation}
\Pr_{T\sim\mathcal{D}_{\mathrm{guard}}}
\left[
U(T,S)<U(T,S_{\mathrm{base}})-\delta
\right]\leq \alpha.
\end{equation}
Here \(U(T,S)\) denotes the reported task utility, such as AUC or accuracy; \(\delta\) is the tolerated degradation margin, and \(\alpha\) controls the allowed probability of harmful intervention. This objective characterizes SkillTFM's boundary expansion criterion: the system expands the operating boundary of the base model only when the current evidence supports a skill and the skill satisfies the no-harm requirement.

The difficulty of this decision lies in the fact that the effectiveness of a repair is not a global property of an operator. The same repair action may improve performance in one evidence region but cause harm in another. SkillTFM therefore considers current evidence, the historical reliability of candidate skills, execution risk, and failure routes recorded in skill memory.

\subsection{Evidence-Conditioned Boundary Expansion}

SkillTFM uses the external skill state to perform boundary expansion on the current task. We write the skill state as
\begin{equation}
S=(\mathcal{E},\mathcal{O},\mathcal{R},
\mathcal{G},\mathcal{F},\mathcal{B},\mathcal{P}),
\end{equation}
where \(\mathcal{E}\) denotes evidence extraction rules, \(\mathcal{O}\) the repair skill bank, \(\mathcal{R}\) ranking rules, \(\mathcal{G}\) risk guards, \(\mathcal{F}\) fallback rules, \(\mathcal{B}\) rejected-route constraints, and \(\mathcal{P}\) principle memory. The lower part of Fig.~\ref{fig:methodology} shows how these components work together on the current task.

First, SkillTFM extracts evidence from the current task and the base-model behavior:
\begin{equation}
z_T=E(T,M).
\end{equation}
As shown in the Evidence Extraction panel, \(z_T\) is used to detect observable signals related to boundary failure, such as distribution change, label-feature relation, prediction bias, trend or periodicity, missingness pattern, and perturbation sensitivity. SkillTFM uses these signals to determine whether the base model exhibits repairable boundary evidence and which repair skills may be applicable. Runtime evidence does not use current test labels or final-heldout feedback.

Given \(z_T\), SkillTFM activates skills in the repair skill bank that are supported by the current evidence. Each repair skill declares its applicability conditions and contraindications. The candidate repair set for the current task is
\begin{equation}
\begin{aligned}
\mathcal{C}(z_T,S)=
\{o\in\mathcal{O}:&
\ z_T\in\mathrm{required}(o),\\
&
\ z_T\notin\mathrm{contraindicated}(o)\}.
\end{aligned}
\end{equation}
Here \(\mathcal{O}\) is the verified repair skill bank, while \(\mathcal{C}(z_T,S)\) contains the candidate repairs activated by the current evidence. The Repair Skills panel in Fig.~\ref{fig:methodology} illustrates this relation: many repair skills may exist in the skill bank, but only those supported by the current evidence and not blocked by contraindications enter the candidate set.

Candidate repairs then enter the gating policy. For each candidate \(o\), SkillTFM computes an auditable ranking tuple:
{\small
\begin{equation}
\begin{aligned}
q(o,z_T;S)=
(&r_{\mathrm{sup}}(o,z_T), r_{\mathrm{hist}}(o;S),
-r_{\mathrm{risk}}(o,z_T;S),\\
&-r_{\mathrm{mem}}(o,z_T;S),-\mathrm{cost}(o)).
\end{aligned}
\end{equation}
}
Here \(r_{\mathrm{sup}}\) measures evidence support, \(r_{\mathrm{hist}}\) summarizes historical reliability, \(r_{\mathrm{risk}}\) estimates execution risk, \(r_{\mathrm{mem}}\) measures similarity to rejected or constrained routes, and \(\mathrm{cost}\) discourages unnecessary intervention. Candidates are lexicographically ranked according to the priority rules stored in the skill state, producing the ranked candidates shown in the figure.

The runtime certificate scans the ranked candidates in order. A candidate is executed only if its evidence is sufficient, its risk is acceptable, and it is not blocked by a previously rejected or constrained route. If the current candidate fails any check, SkillTFM inspects the next ranked candidate; if no candidate passes, the system falls back to the base-model prediction. This corresponds to the Gating Policy in Fig.~\ref{fig:methodology}: evidence must be sufficient, risk must be acceptable, memory constraints must permit the route, and the repair objective must align with the task utility.

If a certified repair exists, SkillTFM returns the repaired prediction. If no candidate is certified, SkillTFM returns the base-model prediction. Both outcomes produce an execution trace that records observed evidence, activated skills, ranked candidates, certificate results, and the action taken, providing evidence for subsequent skill evolution.

Skill memory is consulted as a cross-stage constraint in this process. Rejected routes and principle memory affect candidate activation, ranking, and certification. For example, a repair route that was rejected in a similar evidence region may be assigned lower priority, require stronger evidence, or be prevented from passing the certificate. In this way, SkillTFM incorporates historical failure experience into the current boundary expansion decision.

\subsection{Validation-Gated Skill Evolution}

SkillTFM not only uses existing repair skills to expand the current boundary, but also converts new boundary experience into reusable skills through validation-gated skill evolution. The upper part of Fig.~\ref{fig:methodology} shows this loop: execution traces generate candidate edits, the promotion gate decides whether an edit can become a validated update, accepted edits expand the deployed skill state, and rejected edits enter skill memory as constraints on future candidate proposals and gating decisions.

An execution trace records how the current skill state behaves on a task, including observed evidence, activated skills, ranked candidates, certificate results, and the action taken. A trace provides evidence for later proposals, but does not directly change deployed behavior. SkillTFM allows a candidate edit to become a new executable capability or control rule only after it passes the promotion gate.

A candidate edit is a bounded update to the external skill state. It may include a new repair skill, but it may also revise evidence extraction rules, repair skill definitions, ranking rules, risk guards, fallback rules, rejected-route constraints, or principle memory. Candidate edits may come from local search, failure analysis, accumulated traces, or an LLM-based optimizer. Regardless of source, a candidate edit cannot directly modify the deployed skill state. It must first be organized into a unified and auditable edit format that specifies what is changed, what evidence supports the change, what effect is expected, what risks are involved, and how the edit should be validated.

The promotion gate decides whether a candidate edit enters the deployed skill state. Let \(S_t\) denote the current skill state, and let \(S_t\oplus\Delta S\) denote the candidate state obtained after applying the edit. For an edit \(\Delta S\), SkillTFM measures the intended capability improvement on promotion-selection tasks \(\mathcal{T}_{\mathrm{sel}}\):
\begin{equation}
\begin{aligned}
G_{\mathrm{sel}}(\Delta S)
=
\frac{1}{|\mathcal{T}_{\mathrm{sel}}|}
\sum_{T\in\mathcal{T}_{\mathrm{sel}}}
\big[
U(T,S_t\oplus\Delta S) -U(T,S_t)
\big].
\end{aligned}
\end{equation}
It also measures harm rate on guard tasks \(\mathcal{T}_{\mathrm{guard}}\):
{\small
\begin{equation}
\begin{aligned}
H_{\mathrm{sel}}(\Delta S)
=
\frac{1}{|\mathcal{T}_{\mathrm{guard}}|}
\sum_{T\in\mathcal{T}_{\mathrm{guard}}}
\mathbf{1}\big[
U(T,S_t\oplus\Delta S)<U(T,S_t)-\delta
\big].
\end{aligned}
\end{equation}
}
Here \(\mathcal{T}_{\mathrm{sel}}\) evaluates whether the intended new capability is improved, while \(\mathcal{T}_{\mathrm{guard}}\) checks whether the edit harms existing capabilities or unrelated tasks.

A candidate edit is promoted only if it satisfies all of the following conditions: the intended capability improvement reaches the threshold, the harm rate stays below the allowed limit, regression review passes, and the edit format is valid:
{\small
\begin{equation}
\begin{aligned}
\mathrm{Promote}(\Delta S)=
\mathbf{1}\big[
G_{\mathrm{sel}}(\Delta S)\geq\epsilon,\ 
H_{\mathrm{sel}}(\Delta S)\leq\alpha,\\
\mathrm{RegressionPass}(\Delta S)=1,\ 
\mathrm{SchemaPass}(\Delta S)=1
\big].
\end{aligned}
\end{equation}
}
The state update is
\begin{equation}
S_{t+1}=
\begin{cases}
S_t\oplus\Delta S, & \mathrm{Promote}(\Delta S)=1,\\
S_t, & \mathrm{otherwise}.
\end{cases}
\end{equation}

Accepted edits expand the repairable boundary by adding validated capabilities or control rules to the deployed skill state. Rejected edits are stored as rejected routes and reasons in skill memory, where they constrain future proposals, certification, and promotion. SkillTFM retains the previous validated snapshot for rollback, and final-heldout tasks are used only for final reporting.

\section{Experiments}

In this section, we evaluate SkillTFM around four questions:

(Q1) Can SkillTFM provide training-free adaptation under held-out, mixed, and transferred boundary conditions?

(Q2) Do boundary evidence, gating, and skill evolution explain its improvement and control?

(Q3) Does the learned skill state remain effective across different TFM backbones and optimizer proposers?

(Q4) Does SkillTFM preserve selective repair behavior in real-world electricity-price forecasting?

\subsection{Experimental Setup}

\textbf{Base models.}
We evaluate SkillTFM with TabPFN \cite{hollmann2023tabpfntransformersolvessmall,Hollmann2025}, TabICL \cite{qu2025tabicltabularfoundationmodel}, TabDPT \cite{ma2026tabdptscalingtabularfoundation}, and LimiX \cite{zhang2025limixunleashingstructureddatamodeling}. SkillTFM uses these models as base predictors and performs training-free adaptation through an external skill state rather than parameter updates.

\textbf{Evaluation settings.}
We use controlled boundary suites covering unstable dependencies, distribution shifts, missingness, imbalance, label noise, high-dimensional distractors, and nonlinear structure. Boundary labels are used only for evaluation grouping and are not provided at runtime. We also evaluate a restricted real-world electricity-price forecasting case; raw market records cannot be released, so we report the construction protocol and aggregate results.

\textbf{Metrics.}
For controlled boundary evaluations, we report AUC, AUC improvement, harm rate, and fallback rate over ten random seeds. For electricity-price forecasting, we report MAE, weighted score, active rate, and material harm rate.

\begin{table}[t]
\centering
\tiny
\setlength{\tabcolsep}{3.0pt}
\renewcommand{\arraystretch}{1.03}
\begin{tabular}{lccccc}
\toprule
Condition
& \shortstack{Base\\AUC}
& \shortstack{SkillTFM\\AUC}
& \shortstack{AUC\\Improv.}
& \shortstack{Harm\\Rate}
& \shortstack{Fallback\\Rate} \\
\midrule
Held-out
& $0.644_{\pm 0.019}$
& $0.786_{\pm 0.016}$
& $\mathbf{+0.142_{\pm 0.020}}$
& 0.000
& $44.6\%$ \\

Mixed
& $0.697_{\pm 0.026}$
& $0.832_{\pm 0.022}$
& $\mathbf{+0.135_{\pm 0.027}}$
& 0.000
& $42.1\%$ \\

Generator/severity
& $0.644_{\pm 0.015}$
& $0.772_{\pm 0.011}$
& $\mathbf{+0.128_{\pm 0.015}}$
& 0.000
& $47.7\%$ \\
\bottomrule
\end{tabular}
\caption{
Capability evaluation under boundary shifts. SkillTFM improves all settings with zero observed harm.
}
\label{tab:main-boundary}
\end{table}

\begin{figure}[t]
    \centering
    \includegraphics[width=0.9\linewidth]{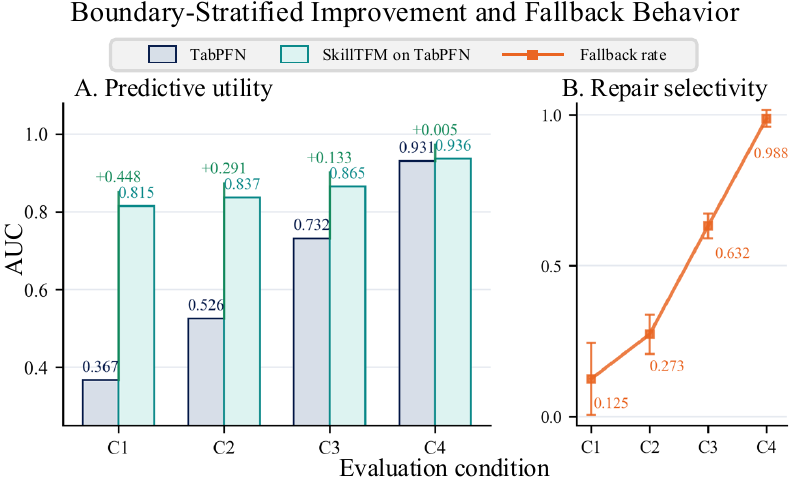}
    \caption{
    Boundary-stratified behavior under mixed-boundary tasks.
    C1--C4 group tasks by base-model AUC ranges
    [0.3,0.4), [0.5,0.6), [0.7,0.8), and [0.9,1.0).
    SkillTFM improves low- and moderate-performance regimes, while the fallback rate increases as the model approaches saturation.
    }
    \label{fig:boundary-stratified}
\end{figure}

\begin{table*}[t]
\centering
\small
\setlength{\tabcolsep}{5pt}
\begin{tabular}{lccccccc}
\toprule
System
& Evolution
& Evidence Retrieval
& Certificate
& AUC
& \shortstack{AUC\\Improv.}
& Harm Rate
& Fallback Rate \\
\midrule

\rowcolor{gray!20}
\multicolumn{8}{l}{\textit{A. Runtime component ablations on the boundary-shift benchmark}} \\
\midrule

Frozen base model
& \xmark & \xmark & \xmark
& 0.672 & 0.000 & -- & -- \\

w/o Evidence Conditioning
& \cmark & \xmark & \cmark
& 0.736 & +0.064 & 0.000 & 63.3\% \\

w/o Runtime Certificate
& \cmark & \cmark & \xmark
& 0.805 & +0.133 & 0.087 & 33.9\% \\

Full SkillTFM
& \cmark & \cmark & \cmark
& 0.817 & +0.145 & 0.000 & 44.8\% \\

\midrule
\rowcolor{gray!20}
\multicolumn{8}{l}{\textit{B. Effect of skill evolution on the nonlinear boundary}} \\
\midrule

Frozen base model
& \xmark & \xmark & \xmark
& 0.699 & 0.000 & -- & -- \\

Static SkillTFM
& \xmark & \cmark & \cmark
& 0.699 & +0.000 & 0.000 & 100\% \\

Full SkillTFM
& \cmark & \cmark & \cmark
& 0.898 & +0.199 & 0.000 & 23.0\% \\
\bottomrule
\end{tabular}
\caption{
Ablation of the core SkillTFM mechanisms.
Panel A evaluates evidence-conditioned retrieval and runtime certification on the boundary-shift benchmark; Panel B evaluates validation-gated skill evolution on the nonlinear boundary.
}
\label{tab:ablation}
\end{table*}

\begin{figure*}[t]
    \centering
    \includegraphics[width=\textwidth]{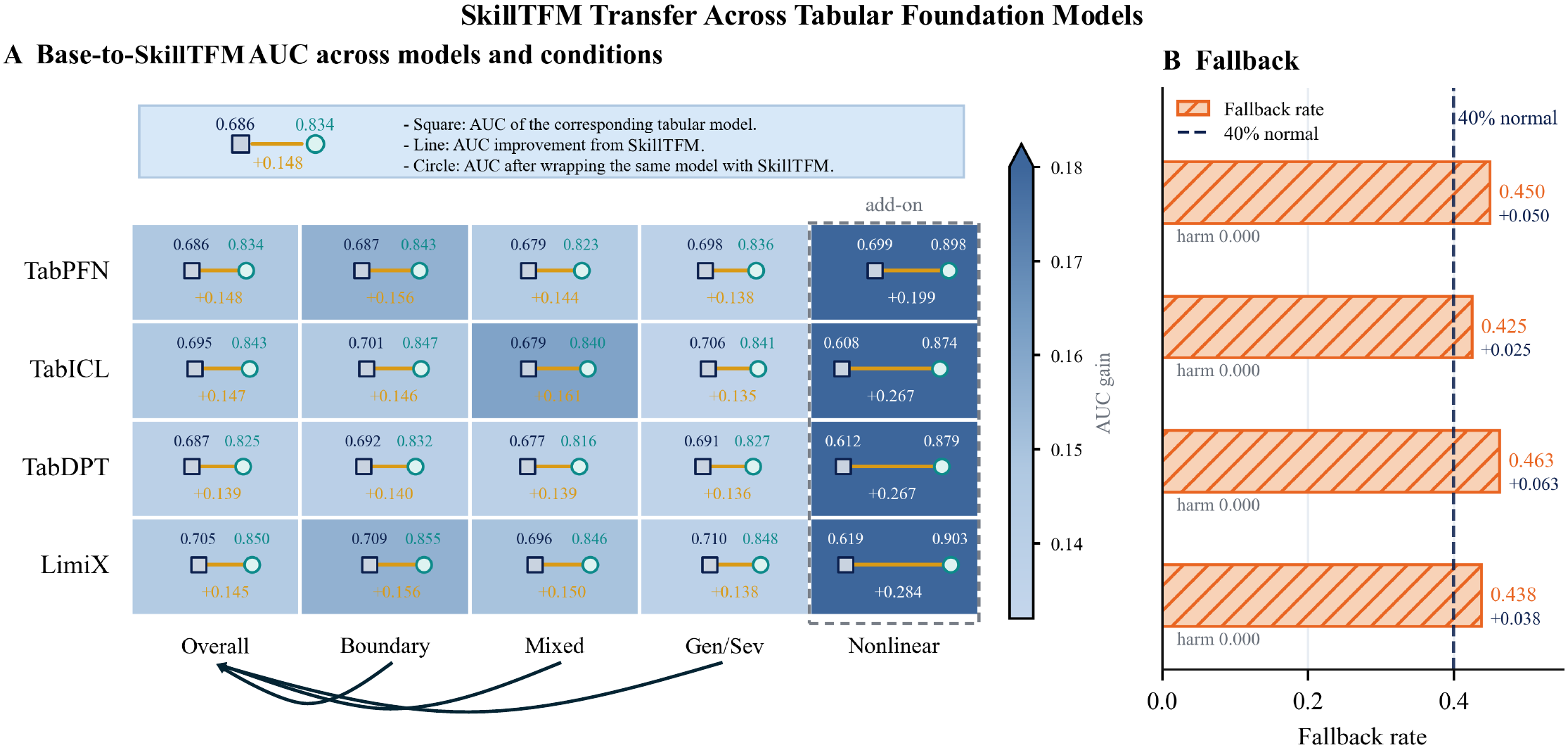}
    \caption{
    Capability extension across frozen foundation models with a shared external skill state.
    Squares denote frozen-model AUC, circles denote AUC after applying the same SkillTFM state, and connecting lines show the resulting improvement.
    The shared state consistently improves all four models while preserving selective intervention and zero observed harm.
    }
    \label{fig:cross-model-transfer}
\end{figure*}

\begin{table*}[t]
\centering
\small
\begin{tabular}{llcccccc}
\toprule
Optimizer & Model & Tokens & Proposed & Valid & Model Val. & Harm & Decision \\
\midrule
Qwen
& Qwen3.7-Max
& 19,404
& 7
& 7
& 1
& 0.000
& Not promoted \\

Gemini
& Gemini-2.5-Flash
& 35,221
& 11
& 11
& 3
& 0.000
& Not promoted \\

Claude
& Claude-Sonnet-4.5
& 31,389
& 11
& 11
& 0
& --
& Evidence insufficient \\
\bottomrule
\end{tabular}
\caption{
Cross-optimizer portability under the shared SkillTFM evolution protocol.
Different LLM optimizers generate schema-compliant edits, while the local promotion gate determines whether they enter the deployed skill state.
}
\label{tab:optimizer-portability}
\end{table*}

\subsection{Q1: Boundary Adaptation and Stratification}

We first evaluate SkillTFM on the main controlled boundary benchmark. The evaluation covers three cases: new tasks with learned single-boundary types, tasks with multiple boundary factors, and tasks where learned boundary types are regenerated with different generators or parameters. Table~\ref{tab:main-boundary} shows that SkillTFM remains effective across all three cases, improving AUC by 0.128--0.142 with zero observed harm. This suggests that the learned skill state is not merely memorizing promotion tasks or applying repairs by default, but using boundary evidence to support gated repair decisions.

We next stratify tasks by base-model AUC to examine when SkillTFM chooses to intervene. Figure~\ref{fig:boundary-stratified} shows that SkillTFM produces the largest improvements when the base model is strongly degraded, but repair is not restricted to the lowest-performing cases: AUC improves by 0.448, 0.291, and 0.133 in C1--C3. In C4, where the base model is already highly reliable, SkillTFM makes almost no change, with an improvement of 0.005 and a fallback rate of 0.988. This pattern shows that the gate is not simply ``repair bad cases and ignore good cases''; it uses boundary evidence to decide whether a repair is supported, including moderate-performance regimes where useful repair is still possible.

\subsection{Q2: Mechanism Ablation and Skill Evolution}

We next examine which mechanisms account for SkillTFM's improvement and selective behavior. Table~\ref{tab:ablation} combines two tests: Panel A ablates the runtime mechanisms used to retrieve and certify repairs, while Panel B evaluates whether validation-gated evolution can add a missing skill for an initially unsupported nonlinear boundary.

In Panel A, removing evidence-conditioned retrieval lowers AUC and increases fallback. The runtime certificate still prevents observed harm, but fewer useful repairs are reached because candidates are no longer selected from current-task boundary evidence. Removing the runtime certificate has the opposite failure mode: more repairs are executed, but harm increases. This shows that evidence-conditioned retrieval mainly controls whether SkillTFM finds actionable repairs, whereas the certificate controls whether those repairs are safe to execute.

Panel B evaluates a capability that is absent from the initial skill state. Before evolution, SkillTFM matches the base model on the nonlinear boundary because it falls back rather than executing an unsupported repair. After validation-gated evolution, a new nonlinear skill is promoted, raising AUC from 0.699 to 0.898 with zero observed harm. The evolution process uses about 107k optimizer tokens, while the base TFM is not retrained. Together, these results show that SkillTFM's training-free adaptation depends on three roles: boundary evidence retrieves candidate skills, gating prevents unsafe execution, and validation-gated evolution expands the skill bank when a new boundary becomes supportable.

\subsection{Q3: Cross-Model Transfer and \\ Optimizer Portability}

We next evaluate whether SkillTFM remains usable beyond the model and optimizer setting in which it is developed. This section tests two forms of portability: whether SkillTFM can transfer across different tabular foundation models, and whether different optimizer proposers can interact with the same skill-evolution interface.

\subsubsection{Cross-Model Transfer}

Figure~\ref{fig:cross-model-transfer} shows that SkillTFM improves TabPFN, TabICL, TabDPT, and LimiX under the same transfer protocol. Across the core transfer benchmark, AUC improvements range from 0.139 to 0.148, indicating that SkillTFM is not tied to a single base-model architecture. The nonlinear add-on column is excluded from the core transfer average, but shows that a newly promoted skill can also be reused across base models.

The fallback analysis shows that this transfer does not come from applying repairs uniformly. The benchmark contains 40\% normal samples and 60\% boundary-affected samples, while fallback rates remain between 0.425 and 0.463 with zero observed harm. Thus, SkillTFM retains selective intervention across base models: it preserves normal or unsupported cases while repairing a large fraction of boundary-affected inputs.

\subsubsection{Cross-Optimizer Portability}

We further evaluate whether skill evolution depends on a particular LLM optimizer. External LLMs act only as candidate-edit proposers under a shared SkillTFM edit interface; all state updates remain controlled by local validation and promotion.

As shown in Table~\ref{tab:optimizer-portability}, Qwen, Gemini, and Claude generate schema-compliant candidate edits under the same interface \cite{qwen2026qwen37max,google2025gemini25flash,anthropic2025claudesonnet45}. Tokens denote the interaction cost required to align each optimizer with the SkillTFM edit protocol, including schema alignment and iterative feedback, rather than the cost of a single proposal. Across optimizers, this cost ranges from 19k to 35k tokens..

No optimizer directly modifies the deployed state. Without additional boundary evidence, no candidate satisfies the promotion criteria. Gemini reaches runtime validation with zero observed harm but insufficient improvement, whereas Claude's candidates are stopped earlier for insufficient evidence. These results separate proposal from admission: optimizers can explore candidate edits, but the local gate decides whether a new capability enters SkillTFM.

\subsection{Q4: Real-World Electricity Forecasting}

We finally evaluate SkillTFM in short-term electricity price forecasting. This case tests whether SkillTFM can extend from controlled tabular boundary tasks to a real temporal prediction setting, where forecasting windows are represented as tabular contexts and repairs may depend on temporal evidence such as periodicity, persistence, and ramp behavior. LimiX serves as the base model and produces 15-minute forecasts for the following three days. Only releases from 2026-06-01 to 2026-06-03 are used for failure analysis, candidate repair search, and skill selection.

To prevent target overlap, the main evaluation reports only the 2026-06-07 predictions from the held-out 2026-06-04 release. The realized 2026-06-07 prices are unavailable in all development releases and are never used for candidate generation, threshold selection, or skill promotion. The test set contains 288 points across three anonymized cities.

\begin{table}[t]
\centering
\scriptsize
\setlength{\tabcolsep}{3.5pt}
\renewcommand{\arraystretch}{1.08}
\begin{tabular}{lrrrrrrr}
\toprule
City
& $N$
& \shortstack{LimiX\\MAE $\downarrow$}
& \shortstack{SkillTFM\\MAE $\downarrow$}
& \shortstack{MAE\\Improv. $\uparrow$}
& \shortstack{Large-error\\Coverage $\uparrow$}
& \shortstack{Active\\Rate}
& \shortstack{Harm\\Rate} \\
\midrule
City A
& 96
& 72.43
& 26.86
& +45.57
& 82.7\%
& 47.9\%
& 0.0\% \\

City B
& 96
& 42.82
& 25.54
& +17.28
& 73.3\%
& 23.5\%
& 0.0\% \\

City C
& 96
& 43.81
& 23.08
& +20.73
& 94.1\%
& 43.3\%
& 0.0\% \\
\midrule
\textbf{All}
& \textbf{288}
& \textbf{53.02}
& \textbf{25.16}
& \textbf{+27.86}
& \textbf{83.4\%}
& \textbf{38.2\%}
& \textbf{0.0\%} \\
\bottomrule
\end{tabular}
\caption{
Electricity price forecasting on the held-out 2026-06-07 targets.
Large-error coverage uses a LimiX absolute-error threshold of 50.
}
\label{tab:electricity-case}
\end{table}

Table~\ref{tab:electricity-case} shows that SkillTFM substantially reduces forecasting error, lowering overall MAE from 53.02 to 25.16. The improvement appears in all three cities, with MAE reductions of 45.57, 17.28, and 20.73.

The repair policy remains selective. SkillTFM modifies 38.2\% of predictions while covering 83.4\% of points where the LimiX absolute error is at least 50, and no activated repair increases absolute error. This case shows that SkillTFM can use learned temporal skills, such as periodic and short-window error-pattern repairs, when a forecasting problem is represented through tabular evidence. It therefore extends the controlled boundary results to a real time-series forecasting setting while preserving gated, non-harmful intervention.

\section{Conclusion}

This paper studies training-free adaptation for tabular foundation models, whose reusable predictions can break under task-specific boundary conditions. We propose SkillTFM, which shifts adaptation from parameter updates to a gated external skill state that selects between evidence-supported repair and fallback. SkillTFM supports validation-gated skill evolution, allowing candidate skills to enter deployment only after utility, risk-control, and regression checks. Experiments across controlled and real-world settings show boundary expansion, demonstrating that SkillTFM expands TFM capability while preserving selective intervention. A current limitation is that uncovered boundary families may still cause failure or harm, and learning such skills requires additional validation data before deployment.

\newpage
\clearpage

\appendix
\bibliography{aaai2027}
\vspace{10em}

\begin{table*}[t]
\centering
\small
\setlength{\tabcolsep}{4.5pt}
\renewcommand{\arraystretch}{1.2}
\begin{tabular}{p{0.17\linewidth}p{0.34\linewidth}p{0.22\linewidth}p{0.20\linewidth}}
\toprule
Boundary family & Construction & Boundary evidence & Intended stress \\
\midrule
Unstable dependency
& The predictive feature relation changes across context and evaluation data.
& Shift in feature-label association or probe sensitivity.
& Reliance on unstable shortcuts. \\

Distribution shift
& Feature marginals or conditional structure are shifted after the context is formed.
& Distributional discrepancy between context and evaluation covariates.
& Deployment mismatch. \\

Missingness
& Structured missing values are injected into informative or correlated features.
& Missingness pattern and feature-level missing rate.
& Sensitivity to incomplete context. \\

Imbalance
& Target or subgroup frequencies are skewed to create rare but important regions.
& Class/subgroup frequency and local support.
& Poor coverage of sparse regions. \\

Label noise
& A subset of context labels is corrupted before prediction.
& Inconsistent label-feature relation or low probe stability.
& Noisy supervision in context. \\

High-dimensional distractors
& Irrelevant or weakly related features are added to the context.
& Feature redundancy, weak association, or dimensional pressure.
& Failure under distracting dimensions. \\

Nonlinear boundary
& The target rule is changed to a nonlinear structure not covered by the initial skill bank.
& Nonlinear probe response and residual structure.
& Need for skill evolution. \\
\bottomrule
\end{tabular}
\caption{Controlled boundary families used to evaluate SkillTFM. Boundary evidence is observable at runtime, while boundary-family labels are used only for evaluation grouping.}
\label{tab:boundary-suite}
\end{table*}

\section{Controlled Boundary Suite Construction}

The controlled suite creates reproducible boundary conditions for tabular foundation models. Each task starts from a clean tabular prediction process, and a boundary mechanism is then injected by modifying the data-generating process, the observed context, or the evaluation distribution. This design tests whether SkillTFM can identify boundary evidence and select an appropriate external skill, rather than only improving on a fixed benchmark distribution.

\paragraph{Clean task process.}
For each task, we first sample a tabular context with mixed informative, weakly informative, and irrelevant features. The target is generated from a latent predictive rule, and the base setting keeps the relation between context and evaluation data stable. Boundary tasks are constructed by perturbing this clean process in controlled ways. Boundary-family labels are used only for suite construction and evaluation grouping.

\paragraph{Boundary families.}
Table~\ref{tab:boundary-suite} summarizes the construction of each boundary family, the observable evidence used by SkillTFM, and the failure mode each family is intended to stress.

\paragraph{Evaluation regimes.}
The same construction protocol forms the controlled evaluation regimes in Table~\ref{tab:main-boundary}: new tasks from learned single-boundary families, tasks with multiple boundary mechanisms, and tasks regenerated from learned boundary families under changed generators or parameters.

\newpage

\begin{algorithm*}[t]
\caption{SkillTFM Boundary Expansion and Validation-Gated Skill Evolution}
\label{alg:skilltfm}
\small
\begin{algorithmic}[1]
\REQUIRE Task \(T\), base TFM \(M\), current skill state \(S_t=(\mathcal{E},\mathcal{O},\mathcal{R},\mathcal{G},\mathcal{F},\mathcal{B},\mathcal{P})\), proposer \(P\), promotion tasks \(\mathcal{T}_{\mathrm{sel}}\), guard tasks \(\mathcal{T}_{\mathrm{guard}}\)
\ENSURE Prediction \(\hat{y}\), updated skill state \(S_{t+1}\), execution trace \(\tau\)

\STATE \textbf{Runtime boundary expansion}
\STATE Compute base prediction \(\hat{y}_M \leftarrow M(T)\)
\STATE Extract boundary evidence \(z_T \leftarrow E(T,M;\mathcal{E})\)
\STATE Initialize candidate set \(\mathcal{C}\leftarrow\varnothing\)

\FOR{repair skill \(o\in\mathcal{O}\)}
    \IF{\(z_T\in\mathrm{required}(o)\) \textbf{and} \(z_T\notin\mathrm{contraindicated}(o)\)}
        \STATE Add \(o\) to \(\mathcal{C}\)
    \ENDIF
\ENDFOR

\STATE Rank candidates in \(\mathcal{C}\) using \(\mathcal{R}\), considering evidence support, historical reliability, execution risk, memory constraints, and cost
\STATE Initialize selected action \(a\leftarrow\mathrm{fallback}\)

\FOR{candidate repair \(o\) in ranked order}
    \STATE Check runtime certificate using \(\mathcal{G}\), \(\mathcal{B}\), and \(\mathcal{P}\)
    \IF{evidence is sufficient \textbf{and} risk is acceptable \textbf{and} memory permits the route}
        \STATE Set \(a\leftarrow o\)
        \STATE Stop scanning remaining candidates
        \STATE \textbf{break}
    \ENDIF
\ENDFOR

\IF{\(a=\mathrm{fallback}\)}
    \STATE Return base prediction \(\hat{y}\leftarrow\hat{y}_M\)
    \STATE Record trace \(\tau\leftarrow(z_T,\mathcal{C},a,\mathrm{fallback})\)
\ELSE
    \STATE Execute certified repair \(\hat{y}\leftarrow a(T,M,\hat{y}_M)\)
    \STATE Record trace \(\tau\leftarrow(z_T,\mathcal{C},a,\mathrm{repair})\)
\ENDIF

\STATE \textbf{Validation-gated skill evolution}
\STATE Generate candidate edit \(\Delta S \leftarrow P(\tau,S_t)\)
\STATE Specify the edit target, supporting evidence, expected effect, possible risk, and validation requirement
\STATE Form candidate state \(S' \leftarrow S_t\oplus\Delta S\)

\STATE Evaluate intended improvement of \(S'\) over \(S_t\) on \(\mathcal{T}_{\mathrm{sel}}\)
\STATE Evaluate harm rate of \(S'\) relative to \(S_t\) on \(\mathcal{T}_{\mathrm{guard}}\)
\STATE Check regression behavior, edit format, and compatibility with the existing skill state

\IF{improvement is sufficient \textbf{and} harm rate is acceptable \textbf{and} regression and format checks pass}
    \STATE Promote edit: \(S_{t+1}\leftarrow S'\)
    \STATE Store \(S_t\) as a rollback snapshot
    \STATE Mark \(\Delta S\) as a validated update
\ELSE
    \STATE Reject edit: \(S_{t+1}\leftarrow S_t\)
    \STATE Add rejected route, rejection reason, and any derived principle to skill memory
\ENDIF

\RETURN \(\hat{y},S_{t+1},\tau\)
\end{algorithmic}
\end{algorithm*}

\section{SkillTFM Pseudocode}
Algorithm~\ref{alg:skilltfm} summarizes both runtime boundary expansion and validation-gated skill evolution. Given a task and a fixed base TFM, SkillTFM extracts boundary evidence, activates and ranks candidate repair skills, and scans them through a runtime certificate; the first certified repair is executed, otherwise the system falls back to the base prediction. The resulting execution trace can then support candidate edits from accumulated traces, failure analysis, local search, or an LLM-based proposer. No edit directly changes the deployed skill state; it must pass promotion checks before becoming a validated update.

\paragraph{Compute environment.}
All controlled experiments were run on an Apple M4 Max workstation with 36 GB unified memory, using CPU execution without GPU acceleration.

\end{document}